\documentclass[10pt,twocolumn,letterpaper]{article}

\usepackage{iccv}
\usepackage{times}
\usepackage{epsfig}
\usepackage{graphicx}
\usepackage{amsmath}
\usepackage{amssymb}
\usepackage{textcomp}
\usepackage{placeins}
\DeclareMathOperator*{\argmin}{\arg\!\min}

\usepackage[table,xcdraw]{xcolor}

\usepackage[pagebackref=true,breaklinks=true,letterpaper=true,colorlinks,bookmarks=false]{hyperref}

\iccvfinalcopy 

\def\iccvPaperID{****} 

\ificcvfinal\fi
\begin{document}

\title{U-mesh: Human Correspondence Matching with Mesh Convolutional Networks}

\author{Benjamin Groisser\\
Technion-Israel Institute of Technology\\
{\tt\small bgroisser@campus.technion.ac.il}
\and
Alon Wolf\\
Technion-Israel Institute of Technology\\
{\tt\small alonw@technion.ac.il}
\and
Ron Kimmel\\
Technion-Israel Institute of Technology\\
{\tt\small ron@cs.technion.ac.il}
}

\maketitle

\begin{abstract}
   The proliferation of 3D scanning technology has driven a need for methods to interpret geometric data, particularly for human subjects. 
   In this paper we propose an elegant fusion of regression (bottom-up) and generative (top-down) methods to fit a parametric template model to raw scan meshes. 
   
   Our first major contribution is an intrinsic convolutional mesh U-net architecture that predicts pointwise correspondence to a template surface. 
   Soft-correspondence is formulated as coordinates in a newly-constructed Cartesian space. Modeling correspondence as Euclidean proximity enables efficient optimization, both for network training and for the next step of the algorithm.
   
   Our second contribution is a generative optimization algorithm that uses the U-net correspondence predictions to guide a parametric Iterative Closest Point registration. By employing pre-trained human surface parametric models we maximally leverage domain-specific prior knowledge.
   
   The pairing of a mesh-convolutional network with generative model fitting enables us to predict correspondence for real human surface scans including occlusions, partialities, and varying genus (\eg from self-contact). We evaluate the proposed method on the FAUST correspondence challenge where we achieve 20\% (33\%) improvement over state of the art methods for inter- (intra-) subject correspondence.
   
\end{abstract}

\section{Introduction}

With the increased availability of high-quality 3D scanning systems (depth cameras, LIDAR, stereo-photogrammetry, etc.) it is necessary to develop suitable techniques to make sense of the massive influx of geometric data. 
In many applications, human subjects are the object of interest in these scans, generating a need for fast and robust methods to interpret nonrigid surfaces. 
Indeed, correspondence matching for human body surfaces has attracted considerable academic attention and proven to be a challenging problem in geometric computer vision. Interest has only grown with the rise of teleconferencing, remote medicine, augmented/virtual reality. 
These practical applications reinforce the need for methods that can cope with real scan data exhibiting measurement noise, occlusions/partialities, and areas of self-contact. 
We tackle each part of this challenging task, leveraging prior information where available and developing novel methods to overcome gaps in current techniques.

\begin{figure}[h]
    \centering
    \includegraphics[width=\linewidth]{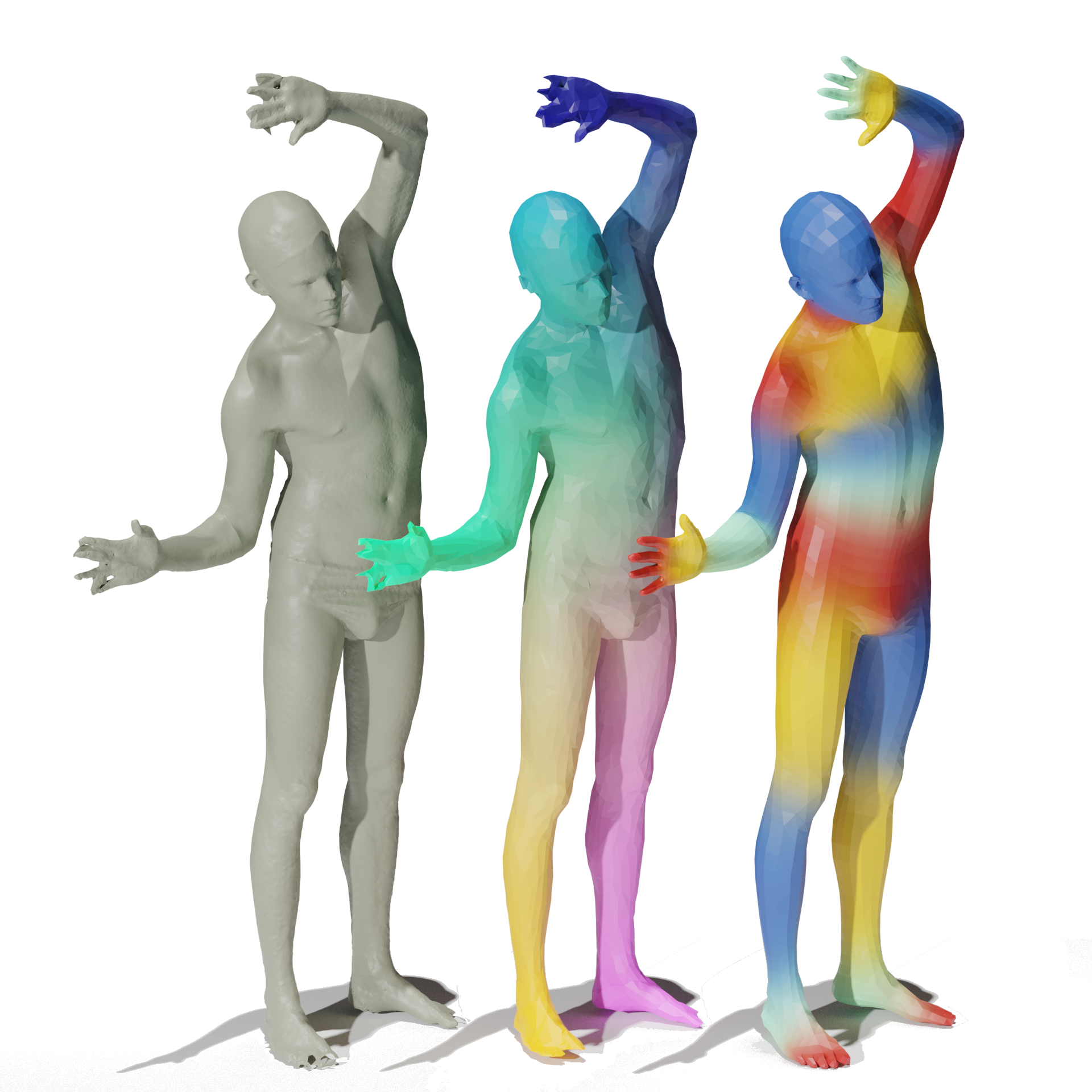}
    \caption{Raw scans (left) are registered to a generative model (right) by way of soft-correspondence predictions (middle).}
    \label{fig:demo}
\end{figure}

\subsection{Regression vs Optimization}

As for so many questions in computer vision, the question immediately arises of how best to leverage prior information (or labelled data) of human body shapes. 
In a 2D context, the question has been posed as a choice between regression (top-down) and generative (bottom-up) approaches \cite{Kolotouros2019a}. 
Should we optimize a parametric model to fit the observed data? 
Or, in the age of deep learning, should we employ a regression model to predict correspondence directly?

In practice, both frameworks are likely necessary; global optimization, though not impossible \cite{Zuffi2015}, is computationally expensive while regression models require enormous quantities of training data and even still typically require post-processing.  
Indeed, state-of-the-art methods use an encoder-decoder network \cite{Groueix2018, Deprelle2019} that naturally combines these approaches. 
At inference time the encoder performs a regression to latent parameters, which provides the initialization for a local optimization to match the output of the decoder to the target surface. 
In our approach we follow a similar scheme but argue that there are compelling advantages to using a purpose-built human body model instead of a generalized decoder.

\subsection{3D Data Structures}

When constructing a regression model for geometric data, we are quickly confronted with the {\it curse of dimensionality}. For 2D image data this problem has been effectively curbed by convolutional neural networks (CNNs) \cite{Fukushima1980, LeCun1989}, as the natural shift-invariance of the domain allows us to re-use model weights and keep the number of trainable parameters manageable. However, non-euclidean domains such as Riemann surfaces do not generally permit a direct analogue of the convolution operation.

Many solutions have been proposed to handle large-scale geometric data; for example, treating point clouds as a set \cite{Qi2016, Zaheer2017} allows a model to flexibly handle data structures of arbitrary size while remaining invariant to ordering. The multi-layer perceptron (MLP) layers of \cite{Qi2016} can be thought of as a single-element convolution, such that the receptive field of each point is restricted to the point itself. 
The authors then sought to overcome the lack of local structure with hierarchical PointNet++ \cite{Qi2017}.

However, when the input data reside in a metric space we have the option to operate in that domain; the surface of a real object (like a human body) natively exists as a 2D manifold embedded in 3D space. 
This naturally invites a surface-based approach, which in practice will commonly be implemented on a triangulated mesh representing the underlying surface.

\section{Contribution}
\label{sec:contribution}

In this paper we present a fully realized pipeline to generate dense correspondence from real human surface scans. We assume that the input is a closed manifold of arbitrary genus, although the techniques described can be readily extended to surfaces with boundaries. 
To introduce this multi-stage algorithm, let us work backward from the desired output.

First, let us note that surface correspondence can equivalently be formulated as a non-rigid registration problem; aligning two surfaces naturally produces correspondence (by proximity) and vice versa. 
We follow the popular framework of mapping all input surfaces to a template shape so that correspondence between input surfaces can be found transitively. 

Our first key intuition is that, contrary to prior surface-matching methods \cite{Groueix2018, Groueix2018a}, there is actually no need to learn a full encoder-decoder model: there already exist powerful human body models \cite{Anguelov2005a, Loper2015a, Osman2020} that span the range of realistic surface shapes. 
These models are low-dimensional, generalize well, decompose identity and pose by construction, and take as input semantic/intelligible parameters that can be constrained and optimized to reproduce a target surface \cite{Hirshberg2012, Saint2019a}. 
Employing a pre-trained model is consistent with our intention to maximally leverage prior information about the possible space of human surfaces.

To bridge the gap between predictive and generative models, we propose an adapted iterative closest point (ICP) algorithm.
This method takes as input initial dense correspondence predictions from a regression model and produces optimized model parameters while aligning a generative model to the original surface. 
Taking a cue from recent regression models \cite{Litany2017} we use a soft-correspondence formulation that can be integrated elegantly into the ICP loop.

The crux of this method, then, is to accurately predict correspondence to a template model from raw surfaces.
To this end we present a novel multi-scale deep convolutional mesh architecture robust to the artifacts commonly exhibited by real scans.
Our mesh operations, including decimation/upsampling and convolutions, can be efficiently implemented to operate at scale, permitting us to train on much larger datasets than previous mesh-based methods.

To take advantage of this multi-scale algorithm, we also present a pipeline to generate realistic synthetic human surface scans with partialities and fused mesh connectivity at areas of self-contact. 
Training with this dataset, in addition to publicly available scans, we significantly outperform the state of the art in human correspondence benchmarks.

\section{Related Efforts}

The general task of non-rigid surface alignment is a core problem of geometric computer vision, and far beyond the scope of this review. 
This discussion will therefore be limited to methods relevant to human correspondence, a rich and varied field by itself. 

\paragraph{Functional Maps}
In contrast to intrinsic formulations, Functional Map methods shift the problem into the frequency domain. Most often, the eigenvectors of the Laplace-Beltrami operator are used as basis functions, as they have been shown to optimally reconstruct smooth functions on surfaces \cite{Aflalo2015}, although recent work has shown that learning task-aware basis functions may improve robustness \cite{Marin2020}. 

Spectral decomposition provides an elegant framework to operate on surfaces, and computational constraints are mitigated by truncating at a low-dimensional basis representation. 
Such methods have proven extremely effective when local deformations are minimal \cite{Litany2017, Halimi2018, Donati2020}; that is, for near-isometries or when a bijective mapping exists. 
However, other classes of transformations (\eg partialities, changes in topological genus, or inter-class comparisons) present challenges that are the subject of ongoing research \cite{Rodola2017, Sharma2021}. 
Furthermore, final correspondence in the spatial domain often requires significant refinement from the initial alignment \cite{Vestner2017, Melzi2019}.

\paragraph{Template Surface Fitting}
As of this writing, the most successful methods for human correspondence matching, according to the FAUST\cite{Bogo2014} correspondence challenge, are built on PointNet-style encoders paired with template-deformation decoder networks \cite{Groueix2018, Deprelle2019}\footnote{Note that many of the methods ranked on the public FAUST benchmark do not represent correspondence matching attempts, and may instead be \eg autoencoders that rely on manual landmarking: \url{http://faust.is.tue.mpg.de/}}. 
These methods preserve local structure by modeling the decoder as deformation of a template surface (with appropriate regularization). This obviates the need for post-hoc filtering but effectively replaces one optimization step with another, as the initial output of the surface autoencoder is generally far from optimal. 
In our method we take inspiration from this fusion of top-down and bottom-up techniques, but argue that better results can be had by swapping out the general deformation model for pre-trained articulated human surface models. 
Furthermore, rather than discarding the local structure of the input surface (\`a la PointNet) we employ an intrinsic mesh-based regression model.

\paragraph{Mesh-Convolutional Architectures}
\label{sec:prior_mesh_conv}

Given the impressive success of CNNs operating on images, it is no surprise that any number attempts have been made to translate these methods to ``geometric'' domains. 
A wide variety of convolution-inspired techniques have been applied to the task of mesh correspondence.

Graph-CNN methods employ the spectral definition of the convolution operator, representing locality as a weight matrix of adjacency between nodes (\eg mesh vertices) \cite{Bruna2014a}. 
Well-defined spatially-localized convolutions can be constructed via smooth spectral multipliers \cite{Henaff2015}, which can be efficiently implemented with recurrent Chebyshev polynomials \cite{Defferrard2016}. 
However, these techniques require a consistent set of basis functions and thus do not translate well to the task at hand, where mesh topology is not consistent.

For spatial formulations, convolution involves an inner product between a kernel function and each operable patch. 
Defining the convolutional neighborhood on a mesh is not trivial, where vertices may have different numbers of neighbors and distances on the surface are non-euclidean.
Masci et al. \cite{Masci2015} described geodesic polar coordinates to align a local region to the convolution kernel.
Subsequent work resolved the orientation ambiguity of the patch operator with anisotropic local coordinates \cite{Boscaini2016}, while other research showed how orientation can be preserved across layers with directional convolutions \cite{Poulenard2018}.

These spatial methods (as well as our work) are part of the {\it template-matching} family of techniques that apply a kernel uniformly across the input domain. 
Other members of the family operate on nearby vertices \cite{Gong2019a, Verma2018}, faces \cite{Milano2020}, or edges \cite{Hanocka2019}. 
In all cases, the challenge is to orient the patches, that is, align the local elements with the convolution kernel. 
Some of these methods have relied on local features \cite{Boscaini2016}, learned attention weights \cite{Monti2017, Verma2018, Milano2020}, or sidestepped the issue by using symmetric convolution filters \cite{Hanocka2019}. 
Our own approach was (very) loosely inspired by the locally flat tangent space mapping of Yang et al. \cite{Yang2018}.

A different strategy is to map the entire surface to a 2D atlas with pre-established coordinates, and perform standard convolutions there. 
Spherical topologies can be parameterized with geometry images, then mapped to a planar domain \cite{Sinha2016}. 
Later studies showed how multiple covering maps of a genus-0 surface allows for a conformal mapping to a flat torus \cite{Maron2017, Haim2019}, a euclidean domain exhibiting translation invariance. 
Taking a data-driven approach, Ezuz et al. train a model to embed the original geometric structure into a euclidean domain \cite{Ezuz2017}. 
Inevitably, any non-isometric mapping involves distortion, the location and extent of which is determined by how the surface is ``cut'' and stretched to the atlas domain.

\paragraph{Human Body Template Models}

Even while interest has exploded for neural networks, significant work has also been done developing generative human body models. 
In the seminal Shape Completion and Animation for PEople (SCAPE) method, Anguelov et al. \cite{Anguelov2005a} showed how a low-dimensional parametric model can faithfully reconstruct an impressive range of realistic surface shapes. 
The basic concept is to model the final surface as the composition of identity and pose, employing a pre-defined kinematic tree to describe articulated joints. 
Body shape is typically expressed using Principal Component Analysis (PCA) \cite{Allen2003}, and data-driven pose dependent deformations are either modeled at the triangle level \cite{Anguelov2005a, Hirshberg2012, Chen2013} or with respect to vertices \cite{Loper2015a, Osman2020, Li2019}. 
In this paper we use the publicly available STAR model \cite{Osman2020} to define our shared template space and to constrain the final registration. 

\section{Proposed Method}

Given a 2D Riemann manifold $\mathcal{X}$ we seek an injective map to a known template surface $\mathcal{T}$. 
Our surface correspondence algorithm consists of two main stages: dense correspondence predictions on the input mesh, which are then used to guide an optimization step to align a template model to the original surface. 

\subsection{Correspondence as Euclidean Proximity}

\begin{figure}
    \centering
    \includegraphics[width=\linewidth]{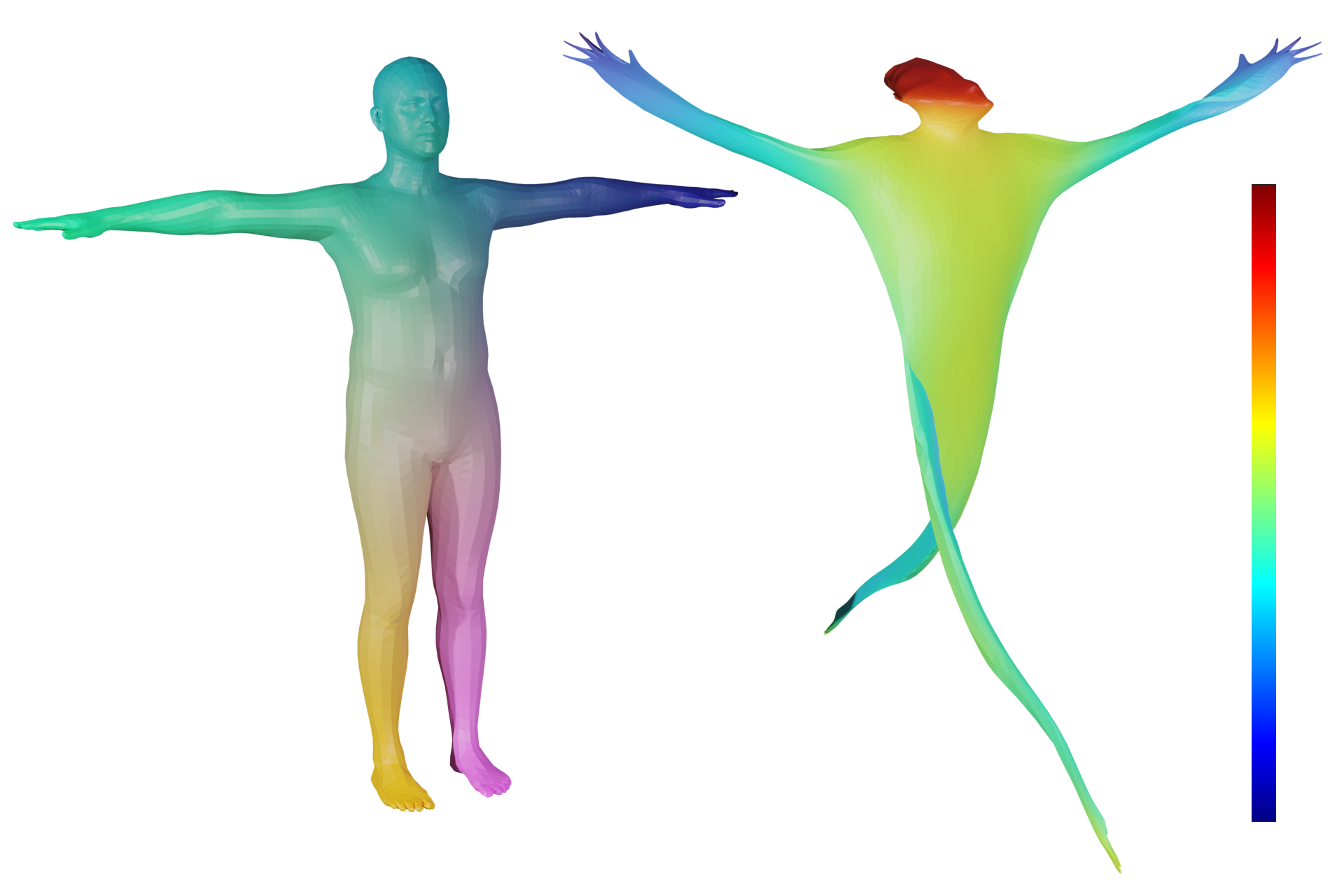}
    \caption{Geodesic soft correspondence is formulated as proximity in a euclidean domain. 
    Coloration of the template model (left) shows the correspondence target as shown throughout the paper. 
    The surface is mapped into a euclidean embedding (right), shown here in $\mathbb{R}^4$ with the fourth dimension represented by color scale.
    }
    \label{fig:template_mds}
\end{figure}

Our conception of correspondence is inspired by the soft error loss proposed by Litany et al. \cite{Litany2017}. 
In their formulation, the loss function was the probabilistic geodesic alignment error between surfaces $\mathcal{X}$ and $\mathcal{Y}$
\begin{eqnarray}
    \ell_F &=& \sum_{(x,y) \in (\mathcal{X}, \mathcal{Y})}{P(x,y)d_{\mathcal{Y}}(y, \pi^*(x))}
\end{eqnarray}
%
where $\pi^*$ is the ground-truth map, $P$ is the pointwise soft correspondence, and $d_{\mathcal{Y}}$ is the geodesic distance on the target surface. 
This error criterion is geometrically well motivated, but suffers in practice at high resolution, as the size of $P$ scales quadratically with mesh sampling. 

We propose a novel approximation of the soft error loss, by projecting the pointwise geodesic distance matrix of $\mathcal{T}$ into a low-dimensional representation using classical multi-dimensional scaling (MDS). 
The result is an embedding of the template surface $\mathcal{T}$ into a coordinate space $\mathbb{R}^d, d \ll n$ where euclidean distances between points approximate the geodesic distances on the original surface. This new euclidean domain offers a compact representation of correspondence to the template; error can be formulated as $\ell_n$ loss, and pointwise matching is performed by a simple nearest-neighbor search. 

It is worth comparing this correspondence error to previous surface-matching techniques. In particular, the use of MDS clearly invites comparison with bending invariant signatures as used for shape classification \cite{Elad2001, EladElbaz2003} or (Spectral) Generalized MDS, in which embeddings are found between surfaces \cite{Bronstein2006, Aflalo2016}. These methods take as inputs near-isometric metric spaces and attempt to perform global alignment, \eg by orthogonal transformation of basis functions \cite{Shtern2014, Shamai2017}. In contrast, our construction treats the euclidean embedding of the template as \emph{absolute coordinates}. The orientation of the target is arbitrary, but it is fixed; rather than aligning global structures or basis function, we aim to predict explicit pointwise correspondence with the template embedding. 

At this point it will be helpful to establish some notation.
\begin{itemize}
    \item $\mathcal{M}_{\xi}$ : real-world $\mathbb{R}^3$ space (Figure \ref{fig:template_mds} left)
    \item $\xi$ : a point in real-world coordinates, $\xi \in \mathcal{M}_{\xi}$
    \item $n$ : normalized normal vector in Cartesian space
    \item $\mathcal{M}_{\omega}$ : MDS euclidean embedding of geodesic distances on the template surface (Figure \ref{fig:template_mds} right)
    \item $\omega$ : a point in euclidean coordinates of the MDS embedding, $\omega \in \mathcal{M}_{\omega}$
\end{itemize}

\subsection{Deep Mesh Architecture}

To learn a surface-to-template correspondence model, we propose a U-net \cite{Ronneberger2015} style architecture operating on $\mathcal{X}$ directly. Cognisant of the challenges inherent to spectral representations, we operate on the explicit surface shape with the aim to learn .  
Multi-scale models propagate information between coarse and fine layers, enabling the model to take advantage of relationships between features at different resolutions \cite{}. 

\begin{figure*}[h]
    \centering
    \includegraphics[width=\textwidth]{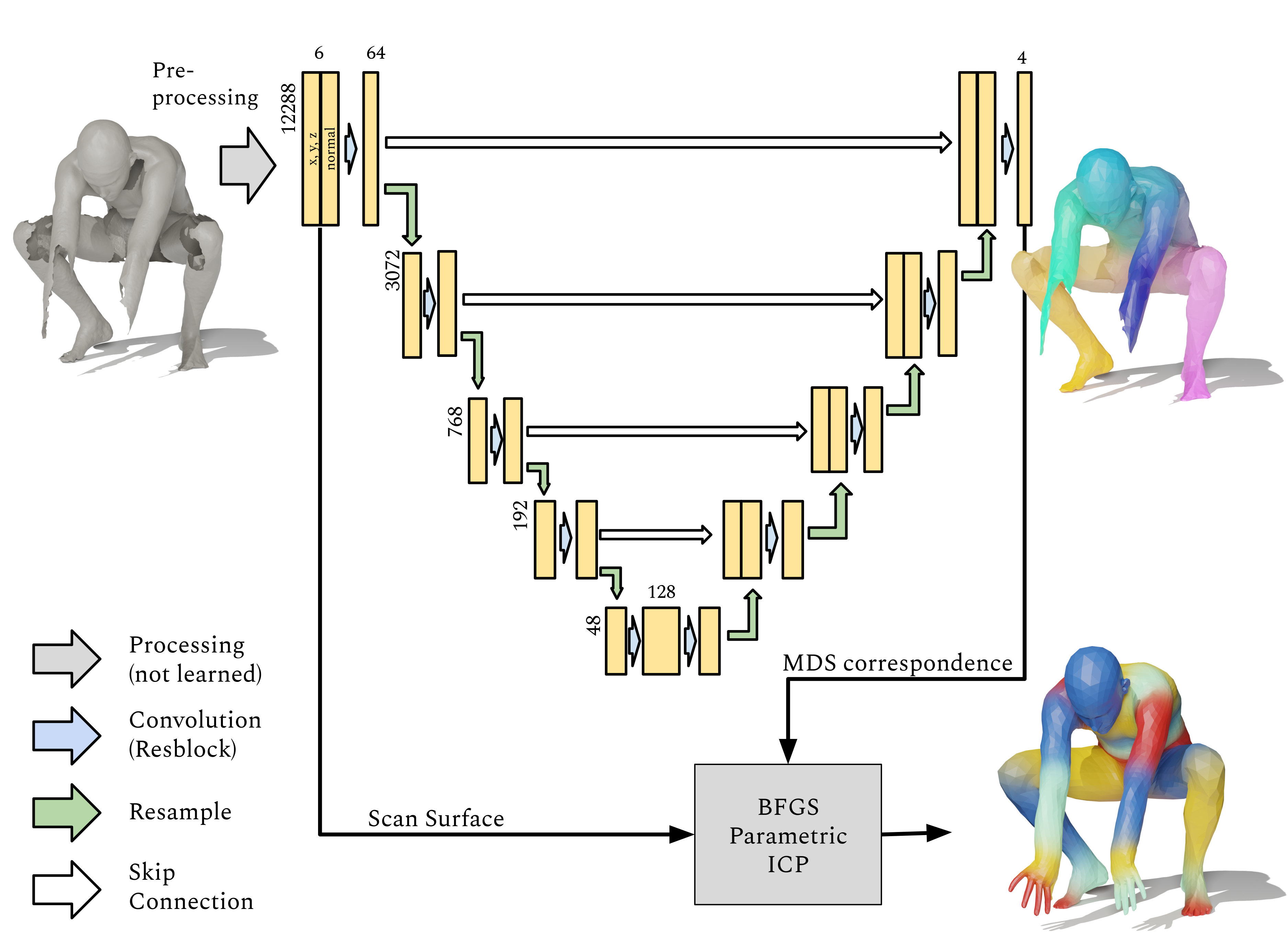}
    \caption{U-mesh architecture. 
    The raw input scan is pre-processed to a standard mesh resolution and all pooling/convolution layers are pre-computed. 
    Feature downsampling is performed by max-pooling, while upsampling uses interpolation. Each convolution is implemented as a resblock \cite{He2016}; unless otherwise indicated, all layers have 64 output features. 
    The scan surface and predicted correspondence are used to optimize model parameters to generate a registered template surface (bottom right). 
    The parametric model enables plausible shape completion in areas of missing data. }
    \label{fig:Umesh}
\end{figure*}

Our goal is to learn an injective function $\Phi : \mathcal{X} \rightarrow \mathcal{T}$, where the output exists in $\mathcal{M}_{\omega}$. 
The (supervised) loss criterion for this model is simply the $\ell_2$ norm of the euclidean distance in $\mathbb{R}^d$, 
\begin{eqnarray}
    \ell_{mds} &=& \sum_{x \in \mathcal{X}} \|\Phi(x) - \Phi^*(x)\|
\end{eqnarray}
where $\Phi^*$ is the ground truth mapping into $\mathcal{M}_{\omega}$. 
Figure \ref{fig:Umesh} shows the complete network architecture; implementation requires that we define surface operations for pooling, up-sampling, and convolutions. 

\subsubsection{Multi-Scale Surfaces}
Surfaces are typically discretized as triangulated mesh structures; this representation is compact and efficiently preserves both the topography and the local geometry of the underlying surface. 
However, even if the exact shape of a surface is known, there are infinite possible samplings resulting in different mesh topologies. 

We pre-process raw meshes by downsampling to a standard mesh size using the Qslim algorithm \cite{Garland1997a}. Quadric error metrics naturally preserve geometric detail, resulting in a mesh structure with dense sampling in high-texture areas and reduced mesh density in flat regions.

Edge-collapse algorithms like Qslim can also be used to generate multi-scale surface samplings, and allow functions defined on the mesh to be propagated between levels. 
In our work we follow a similar implementation to that employed in MeshCNN \cite{Hanocka2019} where edges form the atomic element of mesh operations. 
Starting with a mesh of $m$ edges, mean- or max-pooling can be performed by aggregating the values on incident edges for each edge-collapse. 
Repeated collapses can be tracked sequentially to produce a linear mapping to a new mesh structure of $n$ edges. 

\subsubsection{Mesh Convolutions}

\begin{figure}[htbp]
    \centering
    \includegraphics[width=.8\linewidth, trim={0 6cm 0 4cm},clip]{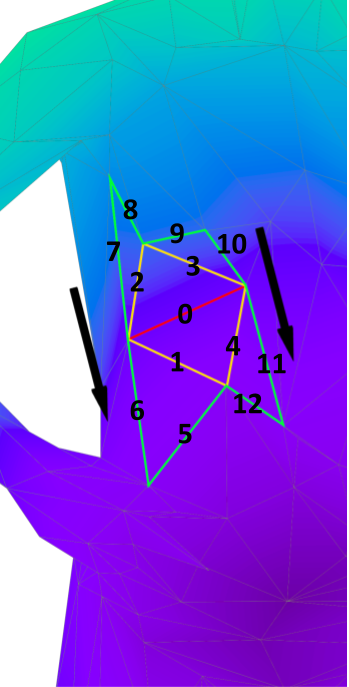}
    \caption{Convolutions are implemented on edges. Here the two-ring neighborhood about an edge is shown along with the gradient of the signal function.}
    \label{fig:conv_neighborhood}
\end{figure}

Closed orientable surfaces do not generally permit a shift-invariant operator. As such, a true analogue to the 2D convolution operation is not possible. 
This points to a related problem: except for an infinitesimally small neighborhood about a point, surfaces do not permit a distortion-free mapping to a euclidean domain. 
Therefore, when defining the region on which a kernel function will be applied we must choose what type of distortion we wish to introduce.

With these considerations in mind, we adopt an entirely intrinsic approach, operating on the \emph{native structure} of the mesh. As described above, we use a geometry-aware decimation technique that preserves surface detail and thereby highlights high-curvature surface structures. 

Convolutions operate on edges as the atomic element in our implementation (Figure \ref{fig:conv_neighborhood}); each edge comprises its own 0-ring and concentric rings are defined by expansion of incident faces. For each patch, the convolution operation is defined as
\begin{equation}
    (f \ast K)(e):= e \cdot k_0 + \sum_{i = 1}^{4}{e_i \cdot k_i} + \sum_{i = 5}^{12}{e_i \cdot k_i} + ...
\end{equation}
where f is a function sampled on the edges, K is the convolutional kernel, and $e_i$ denotes edges at concentric rings about edge $e$. 
This behaviour is well-defined on a manifold triangulated mesh, though overlap is possible.

For patch-based convolutions, the ordering of neighborhood elements is determined by a single degree of freedom, the orientation. 
We propose defining a {\it signal function} over the entire surface, by which all convolution neighborhoods can be oriented. 
In principle any smooth function can be used, but it is desirable to have a global signal that preserves the orientation of nearby patches; this way response to activation functions is coherent across local features. 
In this work we compute the exponential map (geodesic distance) about a single point and orient along the gradient of that function. 
For each surface under consideration the singular point $p_0$ is chosen as the {\it geodesic center of gravity}, i.e.
\begin{eqnarray}
    p_0 &=& \argmin_{p \in \mathcal{X}} \int_{q \in \mathcal{X}}{d(p, q)^2}.
\end{eqnarray}
%
where $d(x,y)$ is the point-to-point geodesic distance across the surface. This construction does result in ambiguity at {\it seams} (where multiple geodesic paths exist), and behavior around $p_0$ is unstable. 
In practice, however, these cases do not pose difficulties.

\subsection{Parametric Model Optimization}
\label{sec:hybrid_icp}

Output from the deep regression model may require refinement; there is no guarantee that the map is injective. 
Furthermore, naively labeling correspondence by nearest neighbor matching may result in extreme mesh deformations \eg if a single point is mapped to a distant location. 
While various refinement algorithms have been described \cite{Vestner2017, Melzi2019}, we opt to rely on a generative model to constrain the final registration.

The final optimization is performed using a form of parametric ICP. 
Our innovation is to perform correspondence matching using our soft correspondence space $\mathcal{M}_{\omega}$, while the error term is computed in real world coordinates $\mathcal{M}_{\xi}$. The error is computed in real-world space $\mathcal{M}_{\xi}$
\begin{eqnarray}
    \ell_{\xi} &=& \frac{1}{a_{\mathcal{X}}} \int_{x \in \mathcal{X}} \| \xi_x - \xi_{t'} \|^2
    \label{eq:icp_dist}
\end{eqnarray}
%
where $a_{\mathcal{X}}$ is the surface area, and $t'$ is the corresponding point on the reconstructed model surface.

Correspondence matching is performed as a weighted combination of nearest neighbor searches in $\mathcal{M}_{\xi}$ and $\mathcal{M}_{\omega}$ in tandem: for each point $x \in \mathcal{X}$ 
\begin{eqnarray}
    t' &=& \argmin_{t \in \mathcal{T}} \| \xi_x - \xi_t \|^2 + \lambda_{\omega} \| \omega_x - \omega_t \|^2 
    \label{eq:correspondence_matching}
\end{eqnarray}
%
where $\lambda_{\omega}$ is a hyperparameter that determines the relative weight given to correspondence predictions vs real-world proximity. 
This {\it guided ICP} can be seen as a sort of continuous version of the Structured Chamfer Distance used in \cite{Li2019}, though we deliberately do not assume a bijective mapping.

\section{Implementation Considerations}

\subsection{Network Architecture}

For correspondence regression, we train a single network for both male and female body types. 
We implement this network in Tensorflow\footnote{Code will be made available after publication}. 
For the target space we use non-metric MDS to embed geodesic distances of the template surface in $\mathbb{R}^d$. 
All experiments share the same euclidean target space: $d = 4$ was chosen after examining the residual  ``strain'' for embeddings of varying dimension, which revealed diminishing returns for higher-dimensional embeddings.

\paragraph{Multi-scale Pre-Processing}

Starting from triangulated meshes of 3D human surfaces, we discard loose patches and repair non-manifold elements before applying Qslim to downsample the input surface to a standard resolution. 
Input features (Cartesian coordinates and normals) are stored in a $m \times 6$ matrix; all experiments use $m_0 = 12288$ edges (8192 faces) as the input size.

Pooling is also performed with Qslim; functions defined on input edges are mapped to subsequent layers via a weighted average\footnote{Each edge collapse directly involves five edges and two vertices, with features defined on edges. Inputs are weighted according to edge length then features on incident edges are averaged to yield feature values of the resulting edges}. 
Tracking these operations from an input mesh of $m$ edges to a decimated mesh of $n$ edges results in a sparse matrix $D \in \mathbb{R}^{n \times m}$ such that mean pooling can be performed with matrix multiplication. 
For max pooling, we threshold $D$ to discard elements less than $0.1$, then use the indices of nonzero elements to determine the receptive field.

\paragraph{Mesh Convolutions}

Patch neighborhoods are defined intrinsically, where for each edge in the mesh a convolution operation maps $k_i$ input features to $k_o$ output features. 
Each edge has two neighbor triangles, each with two additional edges; successive expansions define the N-ring neighborhoods for patch-based operations. 
In our implementation we use a 2-ring neighborhood (Figure \ref{fig:conv_neighborhood}) for a total patch size of 13 edges. 

Convolutions are performed by collecting neighborhood features into a $m \times 13k_i$ matrix. 
The convolution kernel is a $13k_i \times k_o$ matrix, such that convolutions can be performed efficiently by matrix multiplication. 

\subsection{Training}

\subsubsection{Real Data}

To our knowledge, the only large-scale publicly available dataset of human surface scans--including dense labelling--is the DFAUST dataset administered by the Max Plank Institute\cite{Bogo2017}. 
We use the entire ~40k scans in our training set, following the same pre-processing pipeline described above. 

\subsubsection{Synthetic Data}
\begin{figure}[htbp]
    \centering
    \includegraphics[width=\linewidth]{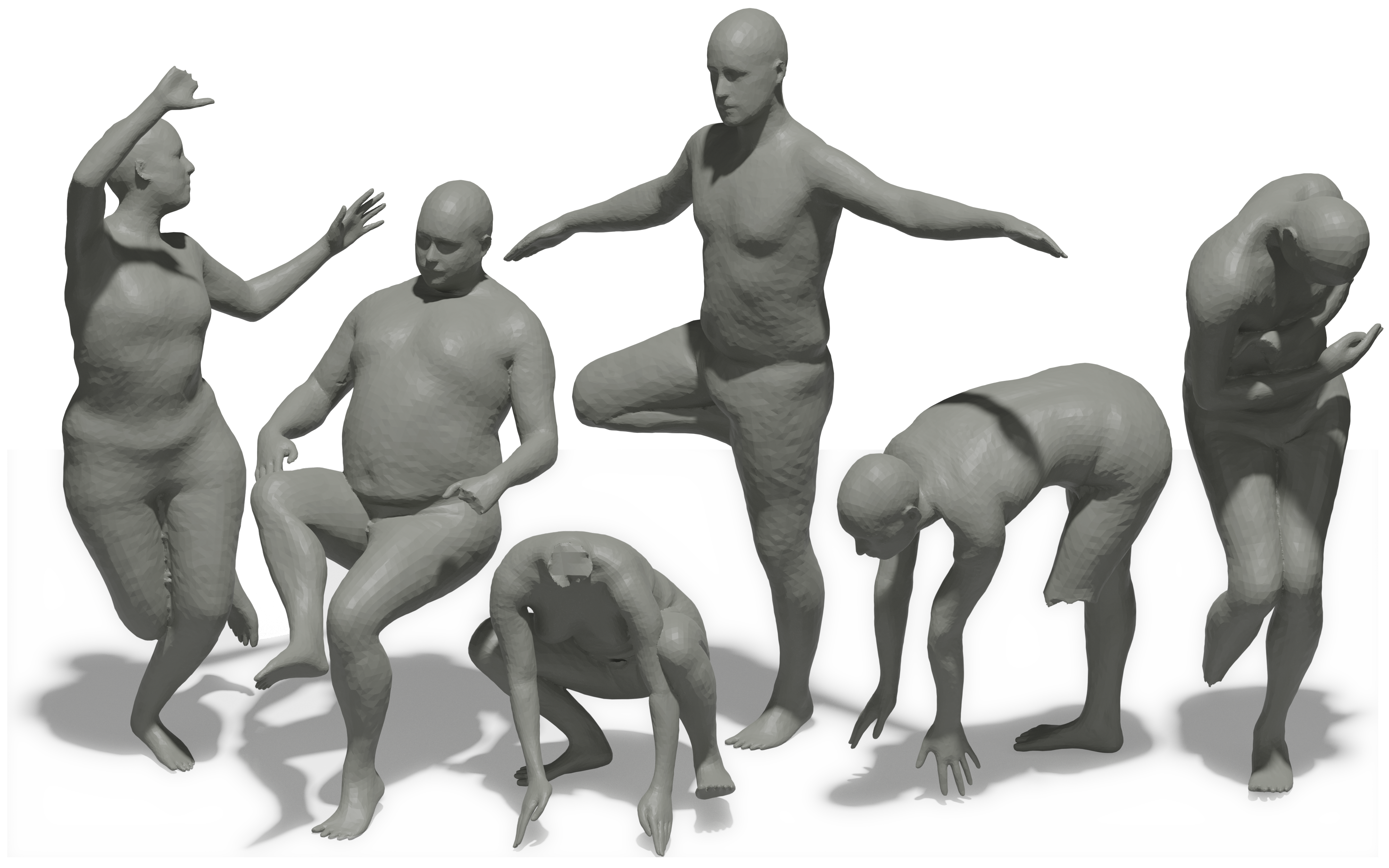}
    \caption{Synthetic scans are generated with the STAR model, then remeshed to fuse the surface at areas of self-contact. Extremities are randomly removed to generate partialities.}
    \label{fig:synthetic}
\end{figure}
The motion sequences captured in DFAUST fall far short of the full breadth of human shape and pose variation. 
To fill this gap we supplement real scan data with synthetic surfaces; we again use the STAR model, this time to produce realistic surfaces with a wide range of human body types.

Approximately 100,000 poses are chosen from the SURREAL \cite{Varol2017} pose library, selecting for maximum variability, which are then used to drive the articulated STAR model. 
We apply ambient occlusion to simulate limited scanner views, then remesh to fuse the surface at contact points. In addition, we stochastically ``amputate'' extremities to generate partialities (Figure \ref{fig:synthetic}).

\subsection{Parametric ICP}

Equation \ref{eq:icp_dist} describes the data term for the model fitting optimization. We regularize the loss function with prior information about the variance of human body shape and joint angles. 
These priors are one of the advantages of using a pre-trained model instead of a general decoder network. 
As is common practice, we penalize the Mahalanobis distance from the expected body shape with
\begin{eqnarray}
    \ell_{\beta} &=& \|\beta - \beta^*\|^2
\end{eqnarray}
%
where $\beta^*$ is the body shape prior, in this case the average (i.e. $\beta^* = \mathbf{0}$). Optimization is performed using the appropriate body model (male/female).

Finally, we regularize relative joint angles with
\begin{eqnarray}
    \ell_{\theta} &=& \|\theta - \theta^*\|^2   ,
\end{eqnarray}
%
where $\theta^* = (\theta_{\max} + \theta_{\min})/2$ is the midrange of joint angles seen in the SURREAL dataset (excluding outliers). 
Parameters $\beta, \theta$ are then optimized using the Tensorflow BFGS solver to minimize
\begin{eqnarray}
    \ell_{ICP} &=& \ell_{\xi} + \lambda_{\beta} \ell_{\beta} + \lambda_{\theta} \ell_{\theta},
\end{eqnarray}
%
where $\lambda_{\beta}$ and $\lambda_{\theta}$ are hyperparameter weighting terms set at $1e^{-3}$ and $1e^{-4}$ respectively. 

Furthermore, when solving \ref{eq:correspondence_matching} we adjust $\lambda_{\omega}$ throughout the model fitting. Initially $\lambda_{\omega} = 20$, favoring the correspondence predictions, gradually reducing until the final convergence depends only on surface matching in $\mathcal{M}_{\xi}$.
Finally, we perform a nonrigid registration to the raw scan data, similar to \cite{Hirshberg2012}; the nonrigid alignment is heavily regularized by the fitted model but can deviate enough to account for irregularities on the observed surface.

\section{Experiments}

\subsection{FAUST}

\begin{figure}[h!]
    \centering
    \includegraphics[width=.98\linewidth]{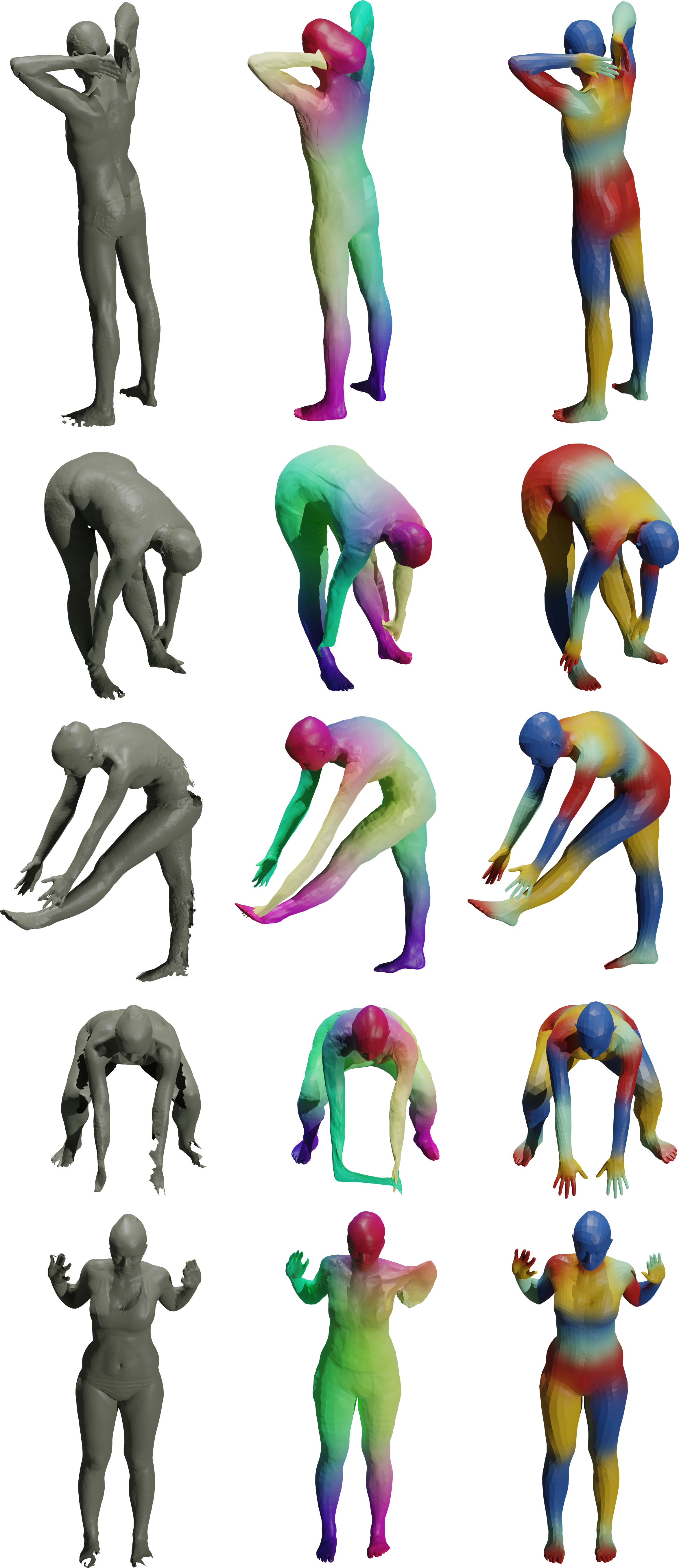}
    \caption{Test scans from the FAUST dataset (left column). The middle column shows results from Deprelle et al. \cite{Deprelle2019}, using their 10d-deformation model, while the right column is our registrations using single-scan optimization. 
    Using a purpose-built template model constrains the solution to the space of realistic human body shapes, and helps account for missing or ambiguous surface structures.}
    \label{fig:registrations}
\end{figure}

\label{sec:faust_results}
The primary benchmark for this method is the FAUST correspondence challenge \cite{Bogo2014}. 
The dataset comprises 300 scans of 10 individual subjects; 100 of those are training samples and include ground truth labelling, while the 200 scans in the test set have labels withheld. 
Two automated benchmarks are available, each computed on a subset of the test scans: 40 scan pairs for the inter-subject and 60 pairs for intra-subject challenge. 
We outperform the state of the art by a significant margin on both benchmarks. 
Processing each scan individually we achieve a 17\% improvement for the inter-subject challenge and a 32\% improvement for the intra-subject challenge.
\begin{table}[h]
    \centering
    \begin{tabular}{|l|l|l|}
    \hline
                                            & Inter             & Intra \\ \hline
    Stitched Puppet \cite{Zuffi2015}        & 3.13              & 1.57  \\ \hline
    FMNet \cite{Litany2017}                 & 4.83              & 2.44  \\ \hline
    3D-CODED \cite{Groueix2018}             & 2.88              & 1.99  \\ \hline
    Elementary Struc. \cite{Deprelle2019}   & 2.58              & 1.63  \\ \hline
    Cyclic FM  \cite{Ginzburg2020a}         & 4.07              & 2.12  \\ \hline
    Ours                                    & \textbf{2.15}     & \textbf{1.07}  \\ \hline
    \end{tabular}
    \caption{Average euclidean error on the FAUST correspondence challenge.}
    \label{tab:comparisons}
\end{table}

\subsection{Decoder}
To examine the importance of the generative optimization step, we experiment with four different decoder algorithms, summarized in Table \ref{tab:decoders}. 
In each case, the same correspondence model (identical U-mesh) is used, so the only difference is in the final registration/refinement.

\paragraph{Single-Scan Correspondence} The baseline algorithm as presented thus far: each test scan is registered independently using Parametric ICP to fit the appropriate STAR model to scan data.

\paragraph{Co-registration} Here we use our prior knowledge of the organizational structure of the test set to fix body shape across all scans for each subject during the final registration. 
Even so we use only the scans in the test set (20 for each subject). 
We argue that this is how the algorithm would presumably be used “in the wild”, where the identity of scan subjects is known and several scans may be captured of the same subject; this assumption is especially appropriate for the intra-subject challenge.
This optimization achieves the highest level of correspondence accuracy, improving on the state of the art by 20\% for the inter-subject challenge and 33\% for the intra-subject subset.

\paragraph{Linear Blend-Skinning} 
For this test we use generic Linear Blend Skinning (LBS) for the generative model. 
This is an articulated mesh model with the same topology and kinematic chain as the STAR model, but without relying on training data--i.e., we use artist-defined blend skinning weights and no pose-dependent deformations. For body shape, 48 parameters are used corresponding to linear scaling of each body segment. 
Performance on the inter-subject benchmark is greatly reduced compared to data-driven template models, but notably we still obtain state-of-the-art results on the intra-subject benchmark. 
This demonstrates that the registered body shapes are subject-consistent, but raw LBS does not generalize well across a population.

\paragraph{Raw U-mesh Correspondence} 
Finally, we perform direct nearest-neighbor matching in $\mathcal{M}_{\omega}$ between challenge pairs. 
Correspondence accuracy is greatly degraded without the generative template registration, showing the importance of refinement and/or model fitting when using regression-based inference.

\begin{table}[]
    \centering
    \begin{tabular}{|l|l|l|}
    \hline
                    & Inter     & Intra \\ \hline
    Single-Scan     & 2.15      & 1.07  \\ \hline
    Coregistration  & 2.06      & 1.06  \\ \hline
    LBS             & 3.45      & 1.52  \\ \hline
    U-mesh          & 4.02      & 3.46  \\ \hline
    \end{tabular}
    \caption{FAUST results using different decoders for final optimization.}
    \label{tab:decoders}
\end{table}

\subsection{Ablation Studies}
We systematically modify the network and/or training inputs and report validation loss on the FAUST challenge.

\subsubsection{Training data}

\begin{figure}[h!]
    \centering
    \includegraphics[width=.98\linewidth]{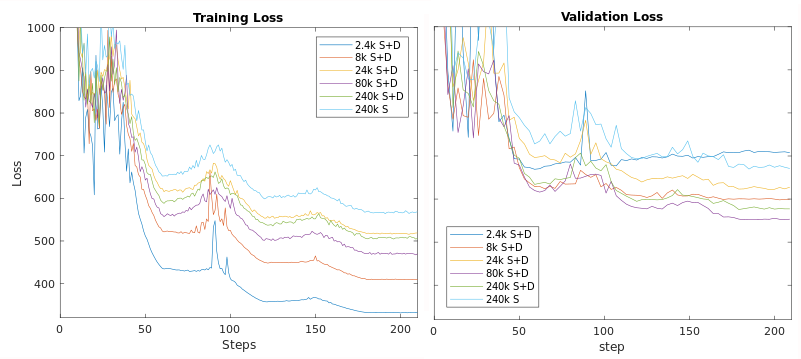}
    \caption{Training and validation loss for different datasets. S indicates synthetic scans, while D represents DFAUST scans used in training.}
    \label{fig:loss_charts}
\end{figure}

We train the U-mesh architecture using different portions of our training data. First, we exclude all DFAUST scans and replace them with additional synthetic scans to see how well our synthetic data generalizes without any real data. Next, we retrain the U-Mesh model using uniformly sampled subsets of the full (synthetic+DFAUST) training set. In all cases, validation losses are computed using the 100 labeled "training" scans from the FAUST dataset. As expected, the model appears to overfit the data with smaller training sets and in the absence of real training scans (Figure \ref{fig:loss_charts}). Remarkably, however, this does not greatly degrade performance in correspondence matching except at severely reduced training set size (Table \ref{tab:training_data}), and even then the final model-fitting optimization almost entirely overcomes that deficit.

\begin{table}[htbp]
    \centering
    \begin{tabular}{l|cc|cc}
              & \multicolumn{2}{c|}{Inter-}  & \multicolumn{2}{c}{Intra-} \\ \cline{2-5} 
              & U-Mesh     & Single-         & U-mesh    & Single-        \\ 
              &            & Scan            &           & Scan           \\ \hline
    240k S    & 4.30       & 2.25            & 3.84      & 1.11           \\
    240k S+D  & 4.02       & 2.15            & 3.46      & 1.07           \\
    80k S+D   & 4.19       & 2.14            & 3.45      & 1.08           \\
    24k S+D   & 4.02       & 2.16            & 3.53      & 1.07           \\
    8k S+D    & 4.11       & 2.14            & 3.72      & 1.07           \\
    2.4k S+D  & 5.28       & 2.17            & 4.25      & 1.19              
    \end{tabular}
    \caption{FAUST correspondence challenge results using different training data. S indicates synthetic scans, while D represents DFAUST scans used in training}
    \label{tab:training_data}
\end{table}

\subsubsection{Convolution ordering}

Defining the convolution neighborhood is the central design choice when designing a patch-based mesh convolution algorithm. 
We have adopted an intrinsic formulation using mesh edges, which affords simplicity of implementation and efficient computations. 
We have previously stated that, in theory, any smooth function defined on the surface may be a candidate to determine neighborhood orientation and we proposed the exponential map about the geodesic center of mass. 
We show the value of this choice by comparing with two naive approaches.
In each case we train using the default 200k synthetic + 40k DFAUST dataset.

First, we train the model with random orientation for each patch; each ring is maintains its ordering, but the indices undergo a cyclic shift. 
This prevents the network from detecting features that are larger than the convolutional kernel, as neighboring patches will no longer share a coherent orientation.
As a result, the network performance is indeed degraded, achieving a raw correspondence error (no generative model) of 5.47cm on the inter-subject FAUST challenge and 4.34cm error for the intra-subject challenge.

For the second test we use an extremely simple ordering scheme, orienting each patch in reference to the vertical axis--i.e. the signal function is simply the height above the floor. In this case we achieve 4.61cm and 4.03cm raw correspondence error on the inter- and intra- subject challenges respectively.

These results, while significantly inferior to the proposed model, are very much in line with performance from a reduced dataset (as above); we conclude that patch orientation is {\it not critically important} to the proposed architecture. 
The multi-scale U-net architecture allows the network to aggregate local information in the downward path and disseminate gross structure in the upward path even without convolutional kernels that act uniformly across local regions.

\subsection{3dBodyTex}

To demonstrate the practical utility of this algorithm, we also perform registrations for the 3dBodyTex dataset \cite{Saint2018}. 
In the original paper they report model-to-scan distances as a measure of the generalizability of registrations. 
This metric is far from ideal however; consider the extreme case where the entire model is collapsed to a single point coincident with a point on the scan surface. 
We therefor report mean distances in both directions (model-to-scan and scan-to-model) for a more complete picture. 
Note that even this does not tell the full story, as there are no guarantees that model and scan anatomy align.

As in \cite{Saint2019a} we compute the model-to-surface distance from each vertex in the model mesh to the closest point on the scan surface, excluding the hands which are sometimes open, sometimes closed in fists, and often include noisy artifacts.
Mean distance across all registrations was $1.5 \pm 9$mm. 
The mean scan-to-model distance was $0.6 \pm 1.7$mm across all vertices in all scans. 
The asymmetry in the error measurements implies that the model is not perfectly aligned anatomically; visual inspection reveals that the full surface of the armpits is often occluded, leaving that region of the model without direct correspondence on the surface.

Our fitting error is lower than the BODYFITR pipeline, but it would be inappropriate to draw any strong conclusions about generalizability from this experiment, since we do not have dense anatomical correspondence as in the FAUST dataset. 
However, the successful surface alignments demonstrate the robustness of the method, at least in the case of relatively clean and complete whole-body scans.
One clear advantage of our method is that it does not require rgb texture, enabling registrations for a wider array of scan systems.

\section{Conclusion}

We have presented a correspondence matching algorithm that combines a deep neural network regression model with a parametric optimization step for robust alignment of real scans with a template model. 
The proposed method provides several tangible benefits over earlier methods. The U-mesh architecture can be trained quickly (several hours) and achieves state of the art performance with only a few thousand training samples. The network requires only the surface geometry as input, making this a useful method even when RGB texture or prior landmarking are not available. 
To construct our deep mesh architecture, we describe several fundamental mesh operations, including geometric pooling/upsampling and an intrinsic edge-based patch convolution. Ordering of convolutional patches is based on a novel {\it signal function}.

To facilitate the transition from pointwise regression to parametric model fitting, we formulate soft geodesic correspondence as  proximity to a fixed template in Euclidean space. 
This enables efficient alignment of a parametric body model with standard gradient descent tools. 

As always, proper training data is crucial; DFAUST scans are augmented with synthetic scan data that are carefully manipulated to reproduce many real-world artifacts: occlusions, partialities, and self contact. The result is a state-of-the-art algorithm for human correspondence matching.

\FloatBarrier


\end{document}